\documentclass[journal,12pt,onecolumn,draftclsnofoot]{IEEEtran}
\usepackage{amsmath}
\usepackage{amssymb}

\usepackage{cite}

\usepackage{algorithm,algorithmic}

\ifCLASSINFOpdf
  \usepackage[pdftex]{graphicx}
\else
\fi
\begin{document}
%
\title{Pareto Deterministic Policy Gradients and Its Application in 5G Massive MIMO Networks}
%
%
%

\author{Zhou Zhou, Yan Xin, Hao Chen, Charlie Zhang, Lingjia Liu
\thanks{Z. Zhou, and L. Liu are with ECE Department at Virginia Tech. Y. Xin, H. Chen and C. Zhang are with Samsung Research America.}
}

%



\maketitle

\begin{abstract}

In this paper, we consider jointly optimizing cell load balance and network throughput via a reinforcement learning (RL) approach, where inter-cell handover (i.e., user association assignment) and massive MIMO antenna tilting are configured as the RL policy to learn. Our rationale behind using RL is to circumvent the challenges of analytically modeling user mobility and network dynamics. To accomplish this joint optimization, we integrate vector rewards into the RL value network and conduct RL action via a separate policy network. We name this method as Pareto deterministic policy gradients (PDPG). It is an actor-critic, model-free and deterministic policy algorithm which can handle the coupling objectives with the following two merits: 1) It solves the optimization via leveraging the degree of freedom of vector reward as opposed to choosing handcrafted scalar-reward; 2) Cross-validation over multiple policies can be significantly reduced. Accordingly, the RL enabled network behaves in a self-organized way: It learns out the underlying user mobility through measurement history to proactively operate handover and antenna tilt without environment assumptions. Our numerical evaluation demonstrates that the introduced RL method outperforms scalar-reward based approaches. Meanwhile, to be self-contained, an ideal static optimization based brute-force search solver is included as a benchmark. The comparison shows that the RL approach performs as well as this ideal strategy, though the former one is constrained with limited environment observations and lower action frequency, whereas the latter ones have full access to the user mobility. The convergence of our introduced approach is also tested under different user mobility environment based on our measurement data from a real scenario.

\end{abstract}

\begin{IEEEkeywords}
    Reinforcement Learning, Multi-Objective Optimization, Online Learning, Load Balancing, FD-MIMO
\end{IEEEkeywords}

\IEEEpeerreviewmaketitle

\section{Introduction}


The global increase of mobile users as well as massive newly-formed wireless connections (e.g. autonomous driving cars and Internet of things) have raised unprecedented challenges to cellular communications systems. Rolling out more intelligent network management approaches is the primary direction beyond the 5th generation cellular system (Beyond 5G) \cite{r2020}. The concept of self-organized network (SON) is summarized in  \cite{klaine2017survey} to address this technique trend, where network techniques are enabled with self-configuration, self-healing and self-optimization 
functionalities. To achieve these goals, the techniques are anticipated to handle the coupling relations between resources allocation, handover management, interference cancellation and coverage optimization, etc.

As a paradigm of the network self-optimization, we consider an automatic way to jointly optimize cell load balance \cite{mao17} and network throughput. Specifically, the realization is required to be as consistent as the operation mode of current cellular systems (i.e. with a smooth evolution). We observe that the inter-cell handover (HO) criteria and massive MIMO (e.g. full-dimension MIMO (FD-MIMO) \cite{nam2013full}) antenna angles are often static in the 4G LTE/LTE advanced system \cite{zhang17}. Motivated by this fact, we consider exploiting the freedom of changing these parameters, i.e. we aim to design an online policy for updating the inter-cell handover thresholds and FD-MIMO antenna tilt angles. More importantly, the optimization objective is no-longer considered as a monopolized metric but with multiple ones.

On the opposite, the non-self-optimization approach is referred to as static optimization approaches where the network parameters are configured according to a pre-assumed model or long-term statistics of the environment. These analytical methods are often unextendable to online fashions due to the bottlenecks of modeling the network dynamics and developing feasible solvers \cite{imran2014challenges}. From the experiments in \cite{ruiz2015analysis}, it demonstrates that directly using the results from a static optimization approach to a dynamic scenario can lead to 85\% to 97\% hand-over failures and connection outages. Also, the negative impacts of model mismatch (applying long-term measurements based strategies to dynamic situations) are discussed in \cite{andrews2014overview}. Accordingly, we can summarize the practical challenges of using these non-adaptive approaches as follows:
\begin{itemize}
    \item Model-based utility optimization cannot extensively incorporate all hidden factors into a unique analytical framework.
    \item Even explicit factors cannot be precisely characterized or timely measured, such as user mobility and the relation between antenna tilt angles and reference signal receiving power (RSRP), etc. 
    \item Usually, a non-convex batch-based optimization problem has to be solved at the end due to the mixed-integer features of network parameters (Here, the antenna tilt angles are discrete variables due to hardware constraints). The non-convex feature makes the static formulation based approaches being intractable online.
\end{itemize}

To adaptively adjust network parameters with light efforts, dynamic programming \cite{ye13} as well as heuristic approaches \cite{damnjanovic2011survey} can be considered. However, these methods often rely on environment assumptions or have uncertainties of their optimality. The randomness of user mobility often results in fluctuated cell-loads and frequent handover \cite{lopez2012mobility} which is the hardest part to characterize. To alleviate any assumptions on the environment, we utilize a reinforcement learning framework (RL) for the online algorithm design. The success of the RL mechanism has been justified in different tasks, such as computer games \cite{schrittwieser2019mastering}: Atari \cite{mnih2013playing} and StarCraft II \cite{vinyals2019grandmaster}, chemical reactions \cite{zhou2017optimizing} and image captioning \cite{mnih2014recurrent} which is our primary motivation of applying RL-based optimization approaches. In our RL approach, a central agent tracks the user measurements and react to favor the predefined objectives to adjust cell individual offsets (CIOs) and BS antenna tilt angles. More importantly, the RL actions are anticipated to perform in a proactive way rather than reactive, i.e., the actions are taken into account of the prediction on user mobility.


In this paper, we introduce a multi-objective RL method and verify the effectiveness of using our measurement data associated with a comprehensive system-level simulator. The main contributions of this paper are as follows,
\begin{itemize}
\item {\textbf{A New Operational Mode:}} Our innovative combination of network handover management and FD-MIMO antenna tilt is a  hierarchical approach for user mobility management: The tilt actions essentially change the cell coverage as a large-scale optimization; The handover management which determines user association is considered as a small-scale adjustment since it especially adheres to the cell edge users. This functional separation is explicitly mapped to the network hardware components which enables enhanced flexibility to the network configuration.

\item {\textbf{A New Algorithm:}} Most existing RL algorithms can only accumulate scalar rewards for a single objective optimization. Whereas, our method can learn through vector rewards thus handle multiple objectives optimization online. To the best of our knowledge, it is the first Pareto-optimization based RL algorithm with continuous action space. Accordingly, this extension allows the underlying policy network to generate mixed actions (i.e., both continuous and discrete actions, where the latter can be easily obtained through quantization). Therefore, the policy on controlling handover threshold (continuous variable) and antenna tilt angles (discrete variable) can be jointly optimized. 

\item {\textbf{Extensive Evaluations:}} We evaluate our method based on a comprehensive simulator which is designed according to 3GPP standards on air interface and access networks\cite{5grrc,5gphysical}. The user side power distribution is obtained from our measurement data in a real city environment. Moreover, we include hardware constraints to the network parameters, such that the FD-MIMO antenna tilt and handover are only allowed to be tuned periodically. Moreover, for self-contained evaluation, we formulate the multi-objective optimization problem through a static perspective. Accordingly, brute-force search based methods are introduced as the benchmark: We give these static methods full access to the user mobility and at the same time relax the periodic action constraints to them. Our results demonstrate that the RL method performs fairly well as the static approaches.
\end{itemize}
In addition, our introduced RL framework can be generalized to handle more than two objectives as well as supporting vector reward with arbitrary dimensions. Also, the action operation can be extended to an asynchronous manner which is considered as a promising direction for our future work.

The rest of this paper is organized as follows: In Section \ref{sec2}, we will review the network optimization methods including both model-based static/online optimization and learning-based approach. In Section \ref{sec3}, we will introduce the background on our network configuration including FD-MIMO antenna tilt, handover management, and network utilities. Section \ref{sec4} will elaborate on our vector reward based multiple objective reinforcement learning methods. Evaluation results are presented in Section \ref{sec6}, where the details of a static solver for the joint optimization will be put in the Appendix as a comparison benchmark. Finally, Section \ref{sec7} will conclude the paper. 

\section{Related Work}
\label{sec2}

Cell load balancing is often considered as the primary objective in network optimization. \cite{ye13} introduced a near-optimal distributed solver in heterogeneous networks. In \cite{bethanabhotla2014user}, an optimizer of user association in massive MIMO systems is proposed. However, the resulting analytical solution is based on a perfect assumption of the massive MIMO channel and without any specific MIMO operations, such as precoding/beamforming. Meanwhile, \cite{razaviyayn2013linear} introduced a downlink precoding method for mobility load balance. But this method does not include the optimization for user association. Moreover, a stochastic geometry based analytical approach is proposed in \cite{singh2013offloading}. While this method can solve a static user association distribution, it cannot be applied online. Meanwhile, it relies on assumptions of user mobility. Overall, the aforementioned approaches are limited to ideal scenarios and with high computational complexity. In current cellular systems, the user association is defined by handover events. Thus the method \cite{Hasan18} containing adaptive optimization on handover parameters by using reference signal receiving power (RSRP) measurements is more relevant. This method is comprised by load prediction and resource adjustment. Indeed, it offers a practical framework for user association optimization. However, it is a heuristic approach, where the optimally cannot be justified. 

Rather than the aforementioned single objective or single variable based optimization, \cite{wang2010dynamic} considers optimizing load balance and maintaining throughput at the same time, where the solver is formulated as an integer optimization. To overcome the computational bottlenecks of this NP-complete problem, online based multiple objective optimizations are proposed in \cite{son2009dynamic,ao2017approximation}. \cite{son2009dynamic} introduced jointly solving the frequency reuse and load balance via changing user association. In addition, the method in \cite{ao2017approximation} considered combining the objectives on throughput maximization and load balance. The result shows that it can achieve $60\%$ - $67\%$ optimality. However, these methods often assume ideal RSRP measurements without handover delay.

Although the previous methods offer certain strategies on network parameter optimization, the performance is vulnerable to environment variation due to model mismatch. Alternatively, learning-based approaches were introduced: \cite{mwanje2013q} proposed a Q-learning based algorithm to adaptively adjust cell individual offsets (CIOs), which is shown to be superior to a CIO adjustment strategy with fixed step-size. 
\cite{mwanje2016cognitive} proposed a mobility load balance algorithm using generalized Q-learning. On the other hand, \cite{kudo2014q} introduced a framework that operates the handover decision merely through user side calculation. Each user is configured with a reinforcement agent which aims to minimize transmission outage and cell load fluctuation. \cite{xu2019deep,xu2019load} introduced using RL for CIO control, where the RL observation is well designed environment features which are demonstrated to be able to effectively handle the curse of dimensionality. Alternatively, an asynchronous multi-agent RL framework is proposed to deal with large number of users in \cite{wang2018handover}. Rather than RL based approaches, \cite{zappone2018user} introduced a feed-forward neural network based supervised learning approach, where the input is directly chosen as the point cloud comprised by user locations and power measurements. Accordingly, the network output is chosen as the ideal user association. However, the operations rely on users' real time feedback to BSs. Meanwhile, predefined training set labels on the optimal user association is required. 

Considering the pros and cons of previous methods, we employ a deterministic poly gradient based RL method \cite{lillicrap2015continuous} to handle the CIO adjustment as well as FD-MIMO antenna tilting. Regarding multi-objective RL techniques, \cite{van2014multi} extended using scalar reward to vector reward based Q-learning approach. However, the action space is restricted as discrete values. In addition, a simple treatment on vector reward by transforming it to scalar reward using scalarization function can be found in \cite{miettinen2002scalarizing}. Nonetheless, this method relies on handcrafted scalarization functions and cross-validation over multiple policies. \cite{van2017hybrid} proposed an RL framework by decomposing a single reward into multiple rewards to accelerate the convergence of Q-learning. Moreover, this method aims at optimization on discrete variables rather than continuous variables.

\section{Network Scenario}
\label{sec3}
\begin{figure}[tp]
    \centering
    \includegraphics[width = 0.6 \linewidth, height = 0.4 \linewidth]{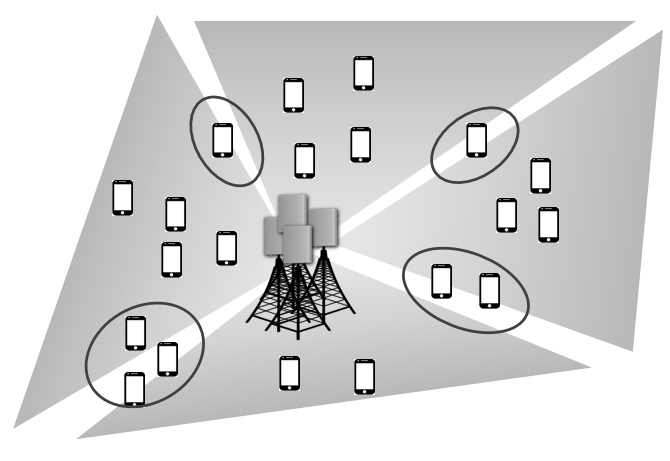}
    \caption{Network layout.}
    \label{network_layout}
\end{figure}

In this section, we will introduce our network layout and utilities. We consider $N$ base stations (BSs) distributed in a city area. For ease of discussion, we denominate the coverage of one BS as a cell, where cells are geographically adjacent to each other, such that Fig. \ref{network_layout} depicts a case of four cells. Our RL agent controls the parameters of the adjacent BSs in a centralized way. We assume there are $K$ users moving around this area. We let $t$ represent the system time index, $k$ be the user index, and $n$ be the BS index. 

\subsection{FD-MIMO with Antenna E-Tilt}
\begin{figure}[tp]
    \centering
    \includegraphics[width = 0.9 \linewidth, height = 0.3 \linewidth]{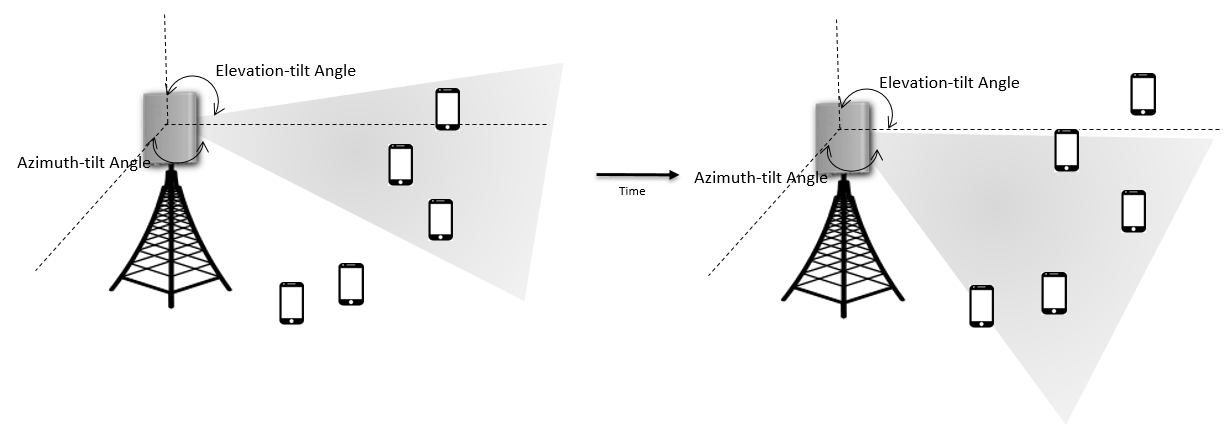}
    \caption{FD-MIMO antenna tilt.}
    \label{antenna_tile}
\end{figure}

For each BS, it is deployed with a 2D antenna array to facilitate the FD-MIMO operation which can offer an extra degree of freedom to the access network. More than operating beamforming towards narrow directions, BS is enabled with an antenna electrical-tilt (E-Tilt) function as shown in Fig. \ref{antenna_tile}, where the antenan tilt contains both elevation and azimuth directions. The tilt action essentially allows a broadcast beam adaption for downlink transmission. Thus, the adjustment of BSs' antenna tilt angles can alternatively change the cell coverage. 
Clearly, the received power by user $k$ from BS $n$ at time instant $t$ is a function of the tilt angle, which we denote as $p_{n,k}(b_n(t))$. $b_n(t)$ comes from a predefined dictionary $\{\theta_0, \theta_1, \cdots, \theta_{M-1} \}$, with $\theta_m$ being a combination of azimuth and elevation angles. For simplicity, we collect all BSs' tilt angles as a vector ${\boldsymbol b}(t)$.\\
{\textbf{ Remarks}} 
\begin{itemize}
\item The downlink received power ${p_{n,k}(b_n(t))}$ can be characterized as a function of the channel realization. However, it is challenging to give an analytical expression of ${p_{n,k}(b_n(t))}$ using both tilt angles and channel coefficients. In our simulation, ${p_{n,k}(b_n(t))}$ is determined via our measurement data set of RSRP. The received signal to noise ratio (SINR) is defined as ${p_{n,k}(b_n(t))\over{\sum_{n' \neq = n} p_{n',k}(b_n(t))}}$, where $n$ is assumed as the serving cell. It is jointly determined by the tilt angles of both serving cell and interference cells. In our system, SINR is first mapped to channel quality indicator (CQI), and then reported to BS in a periodical approach. 
\item Compared to downlink beamforming, the antenna tilt is a slower action due to hardware constraints. Meanwhile, the tilt-action impacts a wider range of users such that adjusts the cell coverage. 
\end{itemize}

\subsection{Handover Management}
The handover process is introduced to switch the user association between cells. Due to user mobility, handover happens when the signal quality from neighbor cells is better than the serving cell. In our system, users report their RSRP measurement to the serving BS, thereafter the BS determines the handover event according to certain criteria. Particularly, we consider A3-event based inter-cell handover which is defined as
\begin{align}
    p_{n',k}(t) - p_{n,k}(t) > O_{n',n}(t) + H_{ys}
\end{align}
where $n$ represents the serving cell and $n'$ represents the neighbor cell; $O_{n',n}(t)$ is the cell individual offset from cell $n$ to $n'$ which is usually symmetrical to $O_{n, n'}(t)$ such that $O_{n',n}(t) = - O_{n, n'}(t)$; $H_{ys}$ is a hysteresis parameter which is set as a constant to avoid frequent handover. When we increase the value of $O_{n, n'}(t)$, the number of handover users can be reduced. Accordingly, the cell load will change, and vice versa. Therefore, adjusting CIO can control the cell load which motivates us to consider it as a parameter in our optimization framework. For ease of discussion, we stack the CIOs as a matrix ${\boldsymbol O}(t)$.

\subsection{Network Utilities}
 Given user SINR and association, we define the following network utilities.
\subsubsection{Cell Load}
We define $l_k(t) := \min \{C_k(t)/r_k(t), l_{limit} \} $ as the load/bandwidth occupation from user $k$ per the unit of physical resource blocks (PRBs), where $r_k(t)$ is the rate of user $k$ which is calculated by mapping through a CQI table adopted in the system;
$C_k(t)$ is the traffic of UE $k$ and $l_{limit}$ is the maximum load allowed for each user; Here, $C_k(t)$ is a pre-given parameter rather than a variable to optimize. For cell $n$, the cell load is given by
\begin{align}
\label{cell_load}
    L_n({\boldsymbol I}(t), {\boldsymbol b}(t)) &= \sum_k I_{n,k} \cdot l_k(t),
\end{align}
where $I_{n,k}$ is the identification for user $k$'s association to BS $n$: $I_{n,k}(t) = 1$ when BS $n$ is associated with user $k$, $I_{n,k} = 0$, otherwise. Note that although we spell out user association rather than handover thresholds as variables of the cell load, CIOs are preferred as the parameters in online optimization due to the feasibility and compatibility. Moreover, the load metric is often normalized by the maximum PRBs per cell.

\subsubsection{Cell Throughput}
Given user association, user rate and user load, we can sum them together to obtain the network throughput,
\begin{align}
\label{throughput}
R_n({\boldsymbol I}(t), {\boldsymbol b}(t)) &= \sum_k I_{n,k} \cdot r_k(t) \cdot l_k(t).
\end{align}
Comparing (\ref{throughput}) to (\ref{cell_load}), we see (\ref{throughput}) is a weighted summation of the same terms in (\ref{cell_load}) which indicates a coupling relation between cell load and throughput.

Intuitively, the coupling relation between cell load balancing and throughput maximization can be observed as follows: We suppose users are \textit{unevenly distributed} in a bandwidth-limited system. To achieve a high throughput (an increase of $\sum_n R_n$), every BS tends to connect to high rate users. Therefore, users are likely to be selected by their nearest BSs. However, this strategy can result in an increase of the peak load in some cells. Consequently, the available PRBs in the high load cells can potentially be lower than the user needed PRBs which will limit the increase of network throughput. Conversely, if BSs hand over some high rate users to neighbor cells to balance the cell load, a high network throughput can potentially be achieved. This is because neighbor cells can allocate more PRBs to low rate users to compensate for the power loss without sacrificing the overall throughput. Thus, choosing proper user association (via CIOs adjustment) and power allocation (through antenna tilt) are important to cell load balancing as well as network throughput maximizing. Moreover, it is important to notice that choosing load balancing as the sole objective is not enough for the optimization. This is because balancing cell load can trigger user link dropping which decreases the overall network throughput. Therefore, the load balancing and throughput maximizing are jointly considered as the objective in this paper. Moreover, when the network is operating with a low peak load, it is more robust to handle user mobility anomalies. For ease of discussion, we summarize the introduced notations in Table \ref{notation_tables}.
\begin{table}[]
\centering
\caption{Notations}
\begin{tabular}{|l|l|l|}
\hline
Symbols & Units &Definition \\
\hline
    $N$  & N/A &The number/set of BSs; Also the set of BSs   \\
\hline
    $K$  & N/A &The number/set of users; Also the set of users \\
\hline
$T$ & h/s &The antenna tilt period; Also the RL action period as well as the time index set\\
\hline
$b_n(t)$ & N/A &The antenna tilt index of cell $n$\\
\hline
$I_{n,k}(t)$ & N/A & Association identification from user $k$ to BS $n$:\\
& &$I_{n,k}(t) = 1$ when BS $n$ is associated with user $k$, $I_{n,k} = 0$, otherwise\\
\hline
 ${\boldsymbol I}(t)$& N/A & User association matrix \\
\hline
${\boldsymbol b}(t)$& N/A & A vector by stacking the tilt index of all BSs \\
\hline
$L_n({\boldsymbol I}(t), {\boldsymbol b}(t))$ or $L_n(t)$ & PRB &The load of cell $n$ \\
\hline
${\boldsymbol l}(t)$& PRB &A vector by stacking the load of all cells: ${\boldsymbol l}(t) = [L_1(t), L_2(t), \cdots, L_N(t)]$\\
\hline
$R_n({\boldsymbol I}(t), {\boldsymbol b}(t))$ or $R_n(t)$  & bit/s &The throughput of cell $n$ \\
\hline
$U({\boldsymbol I}(t), {\boldsymbol b}(t))$ or $U(t)$  & N/A &The tradeoff utility of the overall network \\
\hline
$p_{n,k}(b_n(t))$ or $p_{n,k}(t)$ & mW &The $k$th user's received power from the $n$th cell.\\
\hline
$r_k(t) $ & bit/s/PRB &The rate of user $k$, a function of $p$\\
\hline
$l_k(t)$ & PRB &The load of user $k$, $l_k(t) := \min \{C_k(t)/r_k(t), l_{limit} \} $\\ 
\hline
$l_{limit}$ & PRB &The maximum load constraint for each user\\
\hline
$C_{k}(t)$ & bit/s &The traffic of user $k$, also known as the bit rate requirement\\
\hline
$O_{n, n'}(t)$ & mW &The CIO parameter between cell $n$ and cell $n'$\\
\hline
${\boldsymbol O}(t)$ & mW &The CIO matrix\\
\hline
$e_n(t)$ & N/A &The percentage of cell edge users of cell $n$\\
\hline
${\boldsymbol e}(t)$& N/A &A vector: ${\boldsymbol e}(t) = [e_1(t), e_2(t), \cdots, e_N(t)]$\\
\hline
\end{tabular}
\label{notation_tables}
\end{table}

\section{The Introduced Approach} 
\label{sec4}

In this section, we first introduce some important concepts of reinforcement learning. Then, we will introduce an RL formulation of the online joint optimization between cell loads and throughput. Finally, we will present the details of our algorithm flow. 

\subsection{Preliminary on Reinforcement Learning}
From the perspective of online optimization, reinforcement learning is considered as an approximation to dynamic programming, where the major difference is RL can operate without a mathematical model of the environment. There are two components in the RL framework: an environment and an agent. RL algorithms control the agent to interact with the environment and accumulate rewards from the environment. The RL operation can be generally described as a Markov decision process (MDP) which is a 4-tuple ${\big (}S, A, P_{\boldsymbol a}({\boldsymbol s}(t)|{\boldsymbol s}(t-1), r_{\boldsymbol a}({\boldsymbol s}(t)|{\boldsymbol s}(t-1){\big )}$, where
\begin{itemize}
\item $S$ represent the states which are an interface from the environment to the RL agent.
\item $A$ is the action set at the RL agent. RL agent takes actions to affect the environment, and then receives a change of the states. The action is taken according to a policy denoted as $\pi$.
\item $P_{\boldsymbol a}({\boldsymbol s}(t)|{\boldsymbol s}(t-1))$ represents the station transition probability from ${\boldsymbol s}({t-1}) \in S$ to ${\boldsymbol s}({t}) \in S$ with action ${\boldsymbol a} \in A$.
\item $r_{\boldsymbol a}({\boldsymbol s}(t)|{\boldsymbol s}({t-1}))$ is the immediate reward after action $\boldsymbol a$ is conducted.
\end{itemize}
The Markov property of the states is described via the state transition probability $P_{\boldsymbol a}({\boldsymbol s}(t)|{\boldsymbol s}({t-1}))$. The RL agent takes actions in discrete time steps such that: At time $t$, the agent observes a state ${\boldsymbol s}(t)$. Then, it selects an action ${\boldsymbol a}(t) \in A$ according to a policy $\pi$. The states are updated to ${\boldsymbol s({t+1})}$ associated with sending the RL agent a feed-back reward $r_{{\boldsymbol a}(t)}({\boldsymbol s}(t+1)|{\boldsymbol s}(t))$. Accordingly, this transition can be stacked as a 4-tuple ${\big(}{\boldsymbol s}(t), {\boldsymbol a}(t), r_{{\boldsymbol a}(t)}({\boldsymbol s}(t+1)|{\boldsymbol s}(t)),{\boldsymbol s}(t+1) {\big)}$, namely, a record from the RL experiments. More importantly, the underlying state transition $P_{\boldsymbol a}({\boldsymbol s}(t)|{\boldsymbol s}(t-1))$ and reward $r_{\boldsymbol a}({\boldsymbol s}(t)|{\boldsymbol s}({t-1}))$ cannot be analytically characterized in most scenarios. Thus, RL based approaches fully rely on the samplings from these distributions.

The objective of an RL agent is to accumulate more rewards from the environment in the near future. To achieve this goal, the RL algorithm has two-fold features: 1) It can learn a policy which can offer good received rewards; 2) It can learn a criterion to evaluate whether the policy is good enough. Here, the criterion is named as ``value function'' which estimates the future accumulative rewards as a consequence of using policy $\pi$ from a state $\boldsymbol s$,
\begin{equation}
    \label{value_function}
    V_{\pi}(\boldsymbol s)={\mathbb{E}}_{\pi}\left[\sum_{t=0}^{\infty} \gamma^{t} r_{{\boldsymbol a}(t)}({\boldsymbol s}(t+1)|{\boldsymbol s}(t)) | {\boldsymbol s}({0})={\boldsymbol s}\right],
\end{equation}
where $0\leq \gamma\leq 1$ is a discounted factor introduced to characterize the uncertainties on the future environment and avoid cyclic rewards, and ${\mathbb{E}}_{\pi}$ is taken over the states, rewards and policy, where the policy is usually formulated as a conditional probability: $\pi({\boldsymbol a}(t) |{\boldsymbol s}(t) )$. This is because of the fundamental trade-off between exploration and exploitation in online learning: The exploitation aims to take the best action based on current experiments; The exploration is to gather more information from the environment. Some standard techniques to achieve this trade-off are $\epsilon$ greedy, decayed $\epsilon$ greedy and soft-max, etc. \cite{kaelbling1996reinforcement}.

With a slight modification to the value function, we can define the action-value function \cite{mnih2013playing},
\begin{align}
    \label{Q_definition}
    Q_{\pi}({{\boldsymbol s}, {\boldsymbol a}}) &= {\mathbb{E}}_{\pi}\left[\sum_{t=0}^{\infty} \gamma^{t} r_{{\boldsymbol a}(t)}({\boldsymbol s}(t+1)|{\boldsymbol s}(t)) | {\boldsymbol s}({0})={\boldsymbol s}, {\boldsymbol a}({0})={\boldsymbol a}\right]\\
    &= {\mathbb{E}}_{\pi}\left[ r_{{\boldsymbol a}}({\boldsymbol s}(1)|{\boldsymbol s}) + \sum_{t=1}^{\infty}\gamma^{t} r_{{\boldsymbol a}(t)}({\boldsymbol s}(t+1)|{\boldsymbol s}(t)) \right]\\
    &= {\mathbb{E}}\left[ r_{{\boldsymbol a}}({\boldsymbol s}(1)|{\boldsymbol s}) + \gamma Q_{\pi}({{\boldsymbol s(1)},{\boldsymbol a}({1}})) \right]   
    \label{recursive_form},
\end{align}
where the expectation in (\ref{recursive_form}) operates over the state ${\boldsymbol s}(1)$ and action ${\boldsymbol a}(1)$. We also see that  (\ref{recursive_form}) as a recursive form of the action-value is an unbiased estimation of $Q_{\pi}({{\boldsymbol s}, {\boldsymbol a}})$.  $Q_{\pi}({{\boldsymbol s}, {\boldsymbol a}})$ together with (\ref{recursive_form}) is known as the Bellman equation. In general, when we set the starting time in (\ref{Q_definition}) as $t$, we can immediately attain an expression of $Q_{\pi}({\boldsymbol a}(t), {\boldsymbol s}(t))$. 

Moreover, we can observe that the calculation of the $Q$ function requires the statistics in the future, which is a non-causality formulation. Therefore, to evaluate the current states and actions, we have to first estimate the $Q$ function. This is often achieved by solving the Bellman equation. In many RL algorithms, the solver is based on minimizing a temporal difference (TD) \cite{sutton1988learning} which is defined as
\begin{align}
\label{TD}
    TD(Q) = l(Q_{\pi}({{\boldsymbol s}(t), {\boldsymbol a}(t)}) -  r_{{\boldsymbol a}(t)}({\boldsymbol s}(t+1)|{\boldsymbol s}(t)) - \gamma {\mathbb E}_{\pi'}[Q_{\pi'}'({{\boldsymbol s(t+1)},{\boldsymbol a}({t+1}}))] )
\end{align}
where $l$ represents a predefined loss function, such as ${\mathbb E}|\cdot|^2$, $\pi'$ and $Q'$ are respectively named as target policy and target value function. In this paper, we consider deterministic policy based optimization. Therefore, the inner expectation ${\mathbb{E}}_{\pi'}$ can be drooped. Then, the TD objective becomes
\begin{align}
\label{TD1}
    TD(Q) = l(Q_{\pi}({{\boldsymbol s}(t), {\boldsymbol a}(t)}) -  r_{{\boldsymbol a}(t)}({\boldsymbol s}(t+1)|{\boldsymbol s}(t)) - \gamma Q_{\pi'}'({{\boldsymbol s(t+1)},{\boldsymbol a}({t+1}})) ).
\end{align}
We may further notice that the variable of $TD$ is $Q$ rather than $Q_{\pi}$. This is because we employ an \emph{off-policy} formulation which allows learning the value function and policy in a decoupled way. In general, off-policy can significantly improve data efficiency and offer a fast convergence \cite{munos2016safe}. On the other hand, on-policy learning is often leveraged on $Q_{\pi'}' = Q_{\pi}$ or ${\boldsymbol a}(t+1) \sim \pi$ in (\ref{TD}).

The multiple objective reinforcement learning is defined as the total value function can be decomposed as a linear combination of multiple individual value functions. Such as in the case of two,
\begin{align}
    \label{value_function2}
    V_{\pi}(\boldsymbol s) = w_1 V_{\pi}^{(1)}(\boldsymbol s) + w_2 V_{\pi}^{(2)}(\boldsymbol s)
\end{align}
where $V_{\pi}^{(1)}(\boldsymbol s)$ and $V_{\pi}^{(2)}(\boldsymbol s)$ are two separate value functions following the same definition in (\ref{value_function}); $w_1$ and $w_2$ are weights satisfying $w_1 + w_2 = 1$.

\subsection{RL Formulation}
\label{RL_form}
We follow the routine of MDP to formulate our network parameters optimization. 

\subsubsection{Action Set $A$ }
The action set is considered as antenna tilt and CIOs adjustment as discussed in Section \ref{sec3}, i.e., at time $t$, ${\boldsymbol a}(t) = [{\boldsymbol b}(t), {\boldsymbol O}(t)]$. ${\boldsymbol O}(t)$ and ${\boldsymbol b}(t)$ are characterized as random processes, i.e., $\int P({\boldsymbol O}(t))d{\boldsymbol I}(t) = 1$ and $\int P({\boldsymbol b}(t))d{\boldsymbol b}(t) = 1$. The randomness of ${\boldsymbol O}(t)$ and ${\boldsymbol b}(t)$ is because of the fundamental trade-off on exploitation and exploration of online learning. 

\subsubsection{State Space $S$}
We define the state vector as a stack of the cell loads and the ratio of cell edge users (denoted as ${\boldsymbol e}(t)$), i.e., ${\boldsymbol s}(t) = [{\boldsymbol l}(t), {\boldsymbol e}(t)]$. A user $k$ is counted as an edge user if the throughput is smaller than a pre-defined threshold. This handcrafted state definition essentially captures the features of user geometry: The load reveals the density of users inside the cell, and the edge user ratio reflects the number of potential users for handover events. Comparing with directly using user locations as the states, this reduced state dimension can significantly lower the training cost on the feature extraction. 

\subsubsection{Objective (Value Function)}
For convenience, we first let $\sum_n F(L_n({\boldsymbol I}(t),{\boldsymbol b}(t)))$ (briefly noted as $F(t)$) represent a function which can measure the level of cell load balance. Then, we define the multiple objective optimization as,
\begin{equation}
    \label{optimization0}
    \begin{aligned}
        &\max_{\{{\boldsymbol I}(t): t>0\}, \{{\boldsymbol b}(t): t>0\} } {\mathbb E}[\sum_{t=0}^{\infty}\gamma^t {
            R}(t)] + \lambda{\mathbb E}[\sum_{t=0}^{\infty}\gamma^t {F}(t) ]\\
        &s.t. 
        \quad  \boldsymbol{I}(t)=\{I_{n, k}(t): I_{n,k}(t) \in \{0, 1\}, n \in N, k \in K\}\\
        &\qquad {\boldsymbol b}(t) = \{b_{n}(t):b_{n}(t) \in \{\theta_0, \theta_1, \cdots, \theta_{L-1} \}, n \in N\}\\
        &\qquad {\boldsymbol I}(t) = {\boldsymbol I}(t') \qquad \forall  t, t' \in (pT', (p+1)T'] \quad p \in {\mathbb N} \\
        &\qquad {\boldsymbol b}(t) = {\boldsymbol b}(t')\qquad \forall  t, t' \in (pT, (p+1)T] \quad p \in {\mathbb N}\\
    \end{aligned}
\end{equation}
where the expectation $\mathbb E$ takes into account of all the environmental randomness; $T$ is the time constraint to changing antenna tilt and $T'$ represents the minimum handover time; and $t=0$ stands for the initial operation time.

To be consistent with our action space definition, we have an alternative formulation as follows,
\begin{equation}
\label{optimization1}
\begin{aligned}
    &\max_{\{{\boldsymbol a}(t): t>0\}\sim \pi} {\mathbb E}_{\pi}[\sum_{t=0}^{\infty}\gamma^t {
        R}(t)] + \lambda{\mathbb E}_{\pi}[\sum_{t=0}^{\infty}\gamma^t {F}(t) ]\\
    &s.t. 
    \quad  \boldsymbol{O}(t)=\{O_{n, n'}(t): O_{n,n'}(t)\in [O_{min}, O_{max}], n \in N, n' \in N\}\\
    &\qquad {\boldsymbol b}(t) = \{b_{n}(t):b_{n}(t) \in \{\theta_0, \theta_1, \cdots, \theta_{L-1} \}, n \in N\}\\
    &\qquad {\boldsymbol O}(t) = {\boldsymbol O}(t') \qquad {\boldsymbol b}(t) = {\boldsymbol b}(t')\qquad\\ &\qquad \forall  t, t' \in (pT, (p+1)T] \quad p \in {\mathbb N}\\
\end{aligned}
\end{equation}
Comparing (\ref{optimization0}) to (\ref{optimization1}), we make the following changes,
\begin{itemize}
    \item The variable ${\boldsymbol I}(t)$ is replaced by ${\boldsymbol O}(t)$ which alternatively defines the user association via inter-cell handover. The algorithm is expected to be able to automatically adjust ${\boldsymbol O}(t)$ to proactively change the user association.

    \item The time constraint on CIO is altered to $T$. We make this modification is to synchronize the network adjustment for ease of RL-based action operations. However, the user association is still determined at every $T'$ according to the definition of the A3-handover events.
    
    \item We use ${\boldsymbol a}(t)$ to briefly represent the action policy on ${\boldsymbol O}(t)$ and ${\boldsymbol b}(t)$.
\end{itemize}

\subsubsection{Rewards}

We note that the objective in (\ref{optimization1}) can be merged as ${\mathbb E}[\sum_{t=0}^{\infty}\gamma^t {U}(t)]$, where $U(t) := R(t) + F(t)$. By following the definition of the single objective value function in (\ref{value_function}), we can consider directly set $U(t)$ as the rewards. However, this simple treatment can result in the following ambiguity issue: Suppose we obtain a reward $U$. $U$ can be either composed by $U = R_1 +\lambda F_1$ or $U = R_2  +\lambda F_2$, but $R_1 \neq R_2$ and $F_1 \neq F_2$. This gives us an intuitive understanding that directly merging the two objectives into one and using scalar-reward can potentially cause convergence issues to the RL-based approaches.

To solve the ambiguity issue, we consider defining the reward as a vector: 
\begin{align}
    {\boldsymbol r}_{\boldsymbol a}(t) = [r^{(1)}_{\boldsymbol a}(t), r^{(2)}_{\boldsymbol a}(t)] = [\sum_n R_n(t), \sum_n F_n(t)].
\end{align}
Now, we scale the objective and define ${\boldsymbol w} = [w_1, w_2]$, where $w_1 = {1\over{1+\lambda}}$ and $w_2 = {\lambda\over{1+\lambda}}$. Then, we have the following action-value function,
\begin{align}
    \label{vector_rewards}
    \max_{\{{\boldsymbol a}(t): t>0\}\sim \pi} w_1{\mathbb E}_{\pi}[\sum_{t=0}^{\infty}\gamma^t r^{(1)}_{\boldsymbol a}(t)] + w_2{\mathbb E}_{\pi}[\sum_{t=0}^{\infty}\gamma^t r^{(2)}_{\boldsymbol a}(t) ]\ .
\end{align}
We name the above formulation as a vector reward based multiple objective RL. Intuitively, we can consider $F(t)$ as the peak cell load, i.e., the maximum value of the average cell-load over a period of $T$. Penalizing the peak load can avoid over-loading users in particular cells which alternatively evenly distributes user associations. Accordingly, the reward vector is defined as,
\begin{align}
\label{vector_rewards_1}
{\boldsymbol r}_{\boldsymbol a}(t) = [{1\over T}\sum_{t\in T}\sum_{n\in N} R_n(t), -\max_n {1\over T}\sum_{t\in T}L_n(t)]
\end{align}

Moreover, we can consider a composite term to characterize the cell load balance. For instance, let $F(t) = -\max_n {1\over T}\sum_{t\in T}L_n(t) -\sigma(\{L_n(t)\}_{ n\in N, t\in T})$, where $\sigma$ is the standard deviation function. Accordingly, the two terms of $F(t)$ can be jointly treated as the last entry in the vector reward. But more generally, the second term of $F(t)$ can be included to the vector reward as a new dimension. Therefore, the optimization on the value-action function becomes,
\begin{align}
    \label{vector_rewards2}
    \max_{\{{\boldsymbol a}(t): t>0\}\sim \pi} w_1{\mathbb E}_{\pi}[\sum_{t=0}^{\infty}\gamma^t r^{(1)}_{\boldsymbol a}(t)] + w_2{\mathbb E}_{\pi}[\sum_{t=0}^{\infty}\gamma^t r^{(2)}_{\boldsymbol a}(t) ]+ w_3{\mathbb E}_{\pi}[\sum_{t=0}^{\infty}\gamma^t r^{(3)}_{\boldsymbol a}(t) ]\ .
\end{align}
Thus, the vector reward becomes,
\begin{align}
\label{vector_rewards_2}
{\boldsymbol r}_{\boldsymbol a}(t) = [{1\over T}\sum_{t\in T}\sum_{n\in N} R_n(t), -\max_n {1\over T}\sum_{t\in T}L_n(t), -\sigma(\{L_n(t)\}_{ n\in N, t\in T})]
\end{align}
Note that the above formulation still addresses the same multiple objective optimization defined in (\ref{optimization1}) but using a vector reward with more entries. Therefore, the basic idea of vector reward based RL is to map the features of multiple objectives to multiple dimensions of the vector reward. 

\subsection{Pareto Deterministic Policy Gradient}
\begin{figure}
    \centering
    \includegraphics[width = 0.8 \linewidth, height = 0.5 \linewidth]{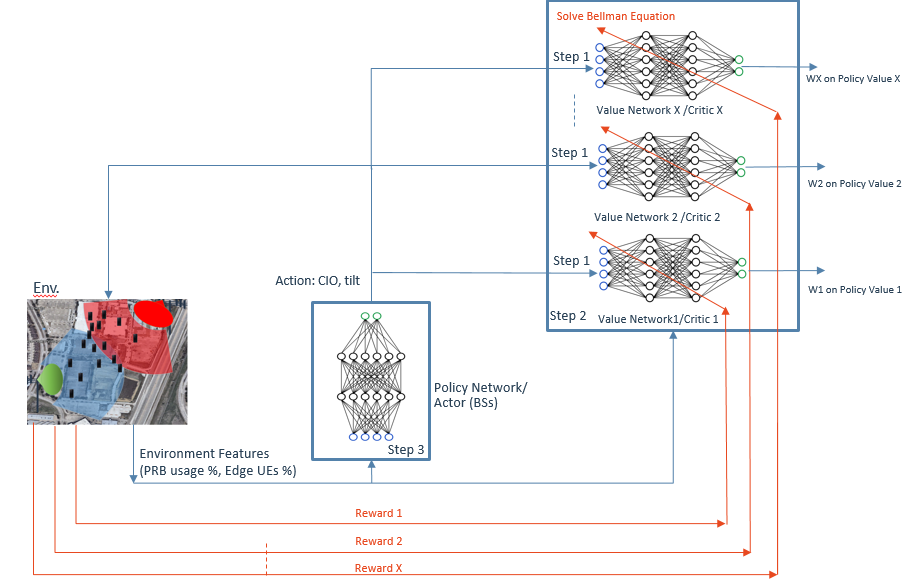}
    \caption{The framework of Pareto Deterministic Policy Gradient.}
    \label{PDPG}
\end{figure}

Now, we consider how to solve vector reward based RL problems. For simplicity, we only elaborate the case of two dimensional rewards, where the reward vector with more than two dimensions can be easily obtained. According to the definition of the action-value function in (\ref{recursive_form}), we can rephrase (\ref{vector_rewards}) as
\begin{equation*}
\label{Q_value}
\begin{aligned}
Q_{\pi}({\boldsymbol s}, {\boldsymbol a}) &= w_1Q_{\pi}^{(1)}({\boldsymbol s}, {\boldsymbol a}) + w_2Q_{\pi}^{(2)}({\boldsymbol s}, {\boldsymbol a})\\
&= w_1{\mathbb{E}}\left[ r_{{\boldsymbol a}}^{(1)}({\boldsymbol s}(1)|{\boldsymbol s}) + \gamma Q_{\pi}^{(1)}({{\boldsymbol s(1)},{\boldsymbol a}({1}})) \right]  + w_2{\mathbb{E}}\left[ r_{{\boldsymbol a}}^{(2)}({\boldsymbol s}(1)|{\boldsymbol s}) + \gamma Q_{\pi}^{(2)}({{\boldsymbol s(1)},{\boldsymbol a}({1}})) \right]
\end{aligned}
\end{equation*}
By rearranging the above equation, we have
\begin{align*}
w_1{\mathbb{E}}\left[(Q_{\pi}^{(1)}({\boldsymbol s}, {\boldsymbol a}) -  r_{{\boldsymbol a}}^{(1)}({\boldsymbol s}(1)|{\boldsymbol s}) - \gamma Q_{\pi}^{(1)}({{\boldsymbol s(1)},{\boldsymbol a}({1}})) \right])
\\= -w_2{\mathbb{E}}\left[(Q_{\pi}^{(2)}({\boldsymbol s}, {\boldsymbol a}) -  r_{{\boldsymbol a}}^{(2)}({\boldsymbol s}(1)|{\boldsymbol s}) - \gamma Q_{\pi}^{(2)}({{\boldsymbol s(1)},{\boldsymbol a}({1}})) \right]).
\end{align*}
This equation implies that the temporal differences on $Q^{(2)}$ and $Q^{(2)}$ are proportional to each other. Motivated by this fact, we define a new TD objective for the vector reward RL as,
\begin{align}
\label{TD_objective}
TD(Q^{(1)}, Q^{(2)}) = w_1 TD^{(1)}(Q^{(1)}) + w_2 TD^{(2)}(Q^{(2)}) .
\end{align} 
where 
\begin{align*}
    TD^{(1)}(Q^{(1)}) = l(Q_{\pi}^{(1)}({\boldsymbol s}(t), {\boldsymbol a}(t))-   r_{{\boldsymbol a}(t)}^{(1)}({\boldsymbol s}(t+1)|{\boldsymbol s}(t)) - \gamma {Q_{\pi'}'}^{(1)}({{\boldsymbol s(t+1)},{\boldsymbol a}({t+1}})))\\
    TD^{(2)}(Q^{(2)}) = l(Q_{\pi}^{(2)}({\boldsymbol s}(t), {\boldsymbol a}(t))- r_{{\boldsymbol a}(t)}^{(2)}({\boldsymbol s}(t+1)|{\boldsymbol s}(t)) - \gamma {Q_{\pi'}'}^{(2)}({{\boldsymbol s(t+1)},{\boldsymbol a}({t+1}}))) .
\end{align*}
By minimizing $TD(Q^{(1)}, Q^{(2)})$, we can simultaneously optimize the temporal difference of $TD^{(1)}(Q^{(1)})$ and $TD^{(2)}(Q^{(2)})$ with weights $w_1$ and $w_2$ respectively.

Moreover, we use neural networks to represent the $Q$ function and policy $\pi$, where they are respectively denoted as ${\mathcal Q}({\boldsymbol s}, {\boldsymbol a}|\theta_{\mathcal Q})$ and ${\mathcal A}({\boldsymbol s}|\theta_{\mathcal A})$. Overall, the learning of $\theta_{\mathcal Q}$ and $\theta_{\mathcal A}$ are briefly summarized as follows: 
\begin{itemize}
    \item Fix $\theta_{\mathcal A}$ and alternatively update $\theta_{{\mathcal Q}^{(1)}}$ and $\theta_{{\mathcal Q}^{(2)}}$ using the gradients of $TD(Q^{(1)}(\theta_{{\mathcal Q}^{(1)}}), Q^{(2)}(\theta_{{\mathcal Q}^{(2)}}))$. 
    \item Fix $\theta_{{\mathcal Q}^{(1)}}$ and $\theta_{{\mathcal Q}^{(2)}}$, and update $\theta_{\mathcal A}$ through the chain rule of $Q({\boldsymbol s}_t, {\mathcal A}({\boldsymbol s}_t|\theta_{\mathcal A}))$.
    \item Repeat the above two stages until convergence.
\end{itemize}
 This alternative updating rule is also called the actor-critic algorithm in the context of RL. As the name suggested, it operates in an actor-critic way: The actor (Policy Network) performs actions and the critic (Value Network) evaluates the action and critiques the actions when low reward values are received. In addition, the output dimension of ${\mathcal A}({\boldsymbol s}|\theta_{\mathcal A})$ is the same as the number of parameters defined in (\ref{optimization0}). Meanwhile, parameters which have integer constraint are quantized afterward. 
 
 Overall, the algorithm is summarized in Algorithm \ref{algorithm1}. We name this method as Pareto deterministic policy gradient algorithm as it optimizes a deterministic policy via using gradient descent according to the action-value objectives. The general framework for the vector reward with more than two dimensions is illustrated in Fig. \ref{PDPG}. We can also notice that our method is based on optimization of a single policy network rather than employing cross-validation over multiple policy networks.
\begin{algorithm}
    \caption{Pareto Deterministic Policy Descent}
    \label{algorithm1}
    \begin{algorithmic}[1]
        \label{algorithm_5}
    \renewcommand{\algorithmicrequire}{\textbf{Input:}}
    \renewcommand{\algorithmicensure}{\textbf{Output:}}
    \REQUIRE Coefficients of policy and actor networks: ${\boldsymbol c}_{\mathcal Q}$, ${\boldsymbol c}_{\mathcal A}$, discounted factor: $\gamma$, soft update parameters: $\tau$, exploration random process: ${\mathcal N}(t)$, normalized tradeoff weights $w = {\lambda \over{1+\lambda}}$
    \ENSURE Online action output: ${\boldsymbol a}(t) = [{\boldsymbol {\tilde I}}(t), {\boldsymbol B}(t)]$
    \STATE \textit{Initialization} : Replay buffer $\mathcal R$; Critic network ${\mathcal Q}_{1}({\boldsymbol s},{\boldsymbol a} |\theta_{{\mathcal Q}_1} = {\boldsymbol c}_{\mathcal Q})$; Critic network 2 ${\mathcal Q}_{2}({\boldsymbol s},{\boldsymbol a} |\theta_{{\mathcal Q}_2} = {\boldsymbol c}_{\mathcal Q})$; Actor network ${\mathcal A} ({\boldsymbol s} |\theta_{\mathcal A} = {\boldsymbol c}_{\mathcal A})$; Target critic networks ${\mathcal Q}'_1({\boldsymbol s},{\boldsymbol a} |\theta_{{\mathcal Q'}_1} ={\boldsymbol c}_{\mathcal Q})$ and ${\mathcal Q}'_2({\boldsymbol s},{\boldsymbol a} |\theta_{{\mathcal Q'}_2} ={\boldsymbol c}_{\mathcal Q})$ and target actor network ${\mathcal A}' ({\boldsymbol s} |\theta_{\mathcal A'}  = {\boldsymbol c}_{\mathcal A})$; $t = 0$
    \FOR{$t$ until the ends}
    \STATE Select action ${\boldsymbol a}(t) = {\mathcal A} ({\boldsymbol s}(t) |\theta_{\mathcal A} ) + {\mathcal N}(t)$
    \STATE Render environment and obtain $r(t)^{(1)}$, $r^{(2)}(t)$ and ${\boldsymbol s}(t+1)$
    \STATE Queue $({\boldsymbol s}(t), {\boldsymbol a}(t), r^{(1)}(t), r^{(2)}(t), {\boldsymbol s}(t+1))$ into a buffer $\mathcal R$
    \STATE Sample a minibatch $\Omega$ from $\mathcal R$
    \STATE Set $\{y^{(1)}(t'): y^{(1)}(t') = r^{(1)}(t') + \gamma Q'_1({\boldsymbol s}(t'+1), {\mathcal A}'({\boldsymbol s}(t'+1)|\theta_{\mathcal A'})|\theta_{\mathcal Q'_1}) , t' \in \Omega\}$ and $\{y^{(2)}(t'): y^{(2)}(t') = r^{(2)}({t'}) + \gamma Q'_2({\boldsymbol s}({t'+1}), {\mathcal A}'({\boldsymbol s}({t'+1})|\theta_{\mathcal A'})|\theta_{\mathcal Q'_2}) , t' \in \Omega\}$
    \STATE Update the two critic networks' coefficients $\theta_{{{\mathcal Q}_1}}$ and $\theta_{{{\mathcal Q}_2}}$ individually by respectively using the gradient $ \sum_{t'}\nabla_{\theta_{{\mathcal Q}_1}}l(y({t'}), {\mathcal Q}_1({\boldsymbol s}({t'}), {\boldsymbol a}({t'})|\theta_{{\mathcal Q}_1}))$ and $ \sum_{t'}\nabla_{\theta_{{\mathcal Q}_2}}l(y({t'}), {\mathcal Q}_2({\boldsymbol s}({t'}), {\boldsymbol a}({t'})|\theta_{{\mathcal Q}_2}))$
    \STATE Update Actor Network $\theta_{\mathcal A} $ using the gradient $w\sum_{t'} \nabla_{{\boldsymbol a}}Q_1(s(t'), {\mathcal A}({\boldsymbol s}(t')|\theta_{\mathcal A})|\theta_{{\mathcal Q}_1}\nabla_{\theta_{\mathcal A}} {\mathcal A}({\boldsymbol s}({t'})|\theta_{\mathcal A}) + (1-w)\sum_{t'} \nabla_{{\boldsymbol a}}Q_2({\boldsymbol s}({t'}), {\mathcal A}({\boldsymbol s}(t')|\theta_{\mathcal A})|\theta_{{\mathcal Q}_2})\nabla_{\theta_{\mathcal A}} {\mathcal A}({\boldsymbol s}({t'})|\theta_{\mathcal A})$
    \STATE 
    \begin{equation}
        \begin{array}{l}
        \theta_{Q'_1}\leftarrow \tau \theta_{Q_1}+(1-\tau) \theta_{{Q^{\prime}_1}} \\
        \theta_{Q'_2}\leftarrow \tau \theta_{Q_2}+(1-\tau) \theta_{{Q^{\prime}_2}} \\
        \theta_{{\mathcal A}^{\prime}} \leftarrow \tau \theta_{\mathcal A}+(1-\tau) \theta_{{\mathcal A}^{\prime}}
        \end{array}
    \end{equation}
    \ENDFOR
    \end{algorithmic} 
\end{algorithm}
In addition, the following experimental techniques are included to stabilize the convergence of our algorithm: 
\begin{itemize}
\item Experiment Replay: We use a replay buffer $\mathcal R$ which is a finite-sized cache to record the transition tuples. The replay buffer is a queue structure that can time out the oldest tuple. At each learning step, networks are updated by sampling a minibatch uniformly from the buffer. This is because PDPG is an off-policy algorithm. Using a large replay buffer can store more uncorrelated transitions thus improve the learning convergence. 
\item Feature Scaling: We scale each entry of the state vector into a certain range and adjust the rewards through shifting and re-scaling. Our experiments prove this technique can make the network learning more effectively. 
\item Noise Exploration: We add an independent Gaussian process with $\sigma$ variance and zero mean to the output of the deterministic policy network to form an exploration policy, i.e.,  ${\boldsymbol a}(t) = {\mathcal A}({\boldsymbol s}(t)) + {\mathcal N}(t)$ . Some other random processes, such as Ornstein-Uhlenbeck process is also widely utilized in the RL literature \cite{lillicrap2015continuous}. Note that the transition history in the experiment replay is based on this exploration policy rather than our target deterministic policy.
\item Soft Target Update: From TD based learning, it is important to choose an updating rule for the target value function and policy to ensure a stable convergence. We follow the “soft” target update strategy, rather than replacing the coefficients of the target networks directly copying from previous steps. The weights $\tau$ is chosen as a very small value to assure the target networks are with a slow change.
\end{itemize}

\section{Evaluation}
\label{sec6}

We evaluate our algorithm on a simulator which is developed based on 3GPP TS-38.331 and 3GPP TS-38.213 \cite{5grrc, 5gphysical}. The simulation environment is a city area with FD-MIMO deployment at BSs. Particularly, we consider applying a RL agent to a $400$m $\times$ $400$m sub-area with 4 BSs. The RL control signal for the 4 cells are assumed synchronized as they are neighboring to each other. In the coverage area of the 4 cells, the average number of users is set as $80$. In the simulator, the user mobility data is generated according to predefined mobility models. Here, we use random way points (RWP) model by default. The model assumes users walking towards random directions with random step sizes. Given users' locations and BS antenna tilt angles, the downlink RSRPs are calculated according to our measurement data which is stored in a look-up table as a 3 mode-tensor, where the three modes are respectively the index of anchor locations, tilt angle and BSs. The number of measurement anchor locations we have in the sub-area is 24,573. The number of tunable antenna tilt angles for each BS is $11$. The maximum time length for the mobility data is set as 200 days.

Given SINR values on the user side, the corresponding CQI values can be obtained. Then the user rate is determined according to a lookup table. In our simulator, the traffic model for each user is set as a constant bit rate (CBR), i.e., $C_k = 1$Mbps. The maximal load for each user is set as 6 PRBs. The total number of PRBs in one cell is 100. The CIO range is [$-12$dB, $12$dB]. When the associated users require more PRBs than the maximal PRBs per cell, the simulator will operate a resource scheduling algorithm afterward, where the eventual PRB allocation is propotional to the rate ranking of scheduled users. 

The parameters in Algorithm \ref{algorithm1} are configured in the intervals $\gamma = [0.1,0.6]$ and $\tau = [0.001,0.01]$. The variance of ${\mathcal N}(t)$ is set as $0.1$ at the beginning of the learning stage to conduct the exploration policy. When the number of iterations is beyond a predefined threshold (our experimental value varies from $100$ to $500$), it is set as a smaller value (such as $100$) to promote the exploitation policy. Both the actor-network and critic-network are chosen as a fully-connected feed-forward neural network with three layers, where the activation functions for the intermediate layers are set as ReLU, the output activation function for the actor neural network is chosen as Tanh function and the critic network output activation is set as linear. The number of neurons for the two intermediate layers are set $50$ and $100$ respectively. For the RL state vector, the throughput threshold for edge user definition is set as 550 kbps. Finally, in the RL algorithm, the period of the antenna tilt and CIO change is set as $2$ hours in terms of the mobility model.

\subsection{Comparison to Different RL Rewards}
\begin{figure}
    \centering
    \includegraphics[width = 0.5 \linewidth, height = 0.4 \linewidth]{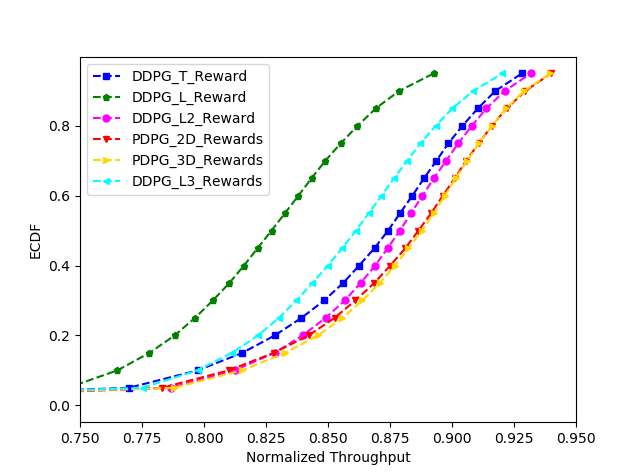}
    \caption{Throughput CDF of using different RL algorithms.}
    \label{Throughput_CDF_RL_Alg}
\end{figure}

We first evaluate the performance using different rewards in the RL algorithm. We consider the CDF curves of network throughput samples from every 15 mins in 200 days amid operating the algorithms to the environment. As illustrated in Fig. \ref{Throughput_CDF_RL_Alg}, the x-axis is chosen as normalized network throughput, where the denominator is chosen as 100Mbps. The labels on the legend respectively represent: DDPG \cite{lillicrap2015continuous} algorithm using scalar reward ${1\over T}\sum_{t\in T}\sum_{n\in N} R_n(t)$, DDPG algorithm using scalar reward $-\max_n {1\over T}\sum_{t\in T}L_n(t)$, DDPG algorithm using scalar reward $w_1{1\over T}\sum_{t\in T}\sum_{n\in N} R_n(t) - w_2\max_n {1\over T}\sum_{t\in T}L_n(t)$, our introduced PDPG algorithm using vector reward (\ref{vector_rewards_1}), PDPG algorithm using vector reward (\ref{vector_rewards_2}) and DDPG algorithm using scalar reward $w_1{1\over T}\sum_{t\in T}\sum_{n\in N} R_n(t) - w_2\max_n {1\over T}\sum_{t\in T}L_n(t) -w_3\sigma(\{L_n(t)\}_{ n\in N, t\in T})$. In this figure, all the weights inside the rewards are well cross-validated to optimize the throughput distribution. In our experience, the vector reward based methods have no much difference on the throughput distributions under different combinations of $w_1$, $w_2$ and $w_3$. Whereas, scalar reward based approaches require more cross-validation tests to obtain the resulting distribution. 

\begin{figure}
    \centering
    \includegraphics[width = 0.5 \linewidth, height = 0.4 \linewidth]{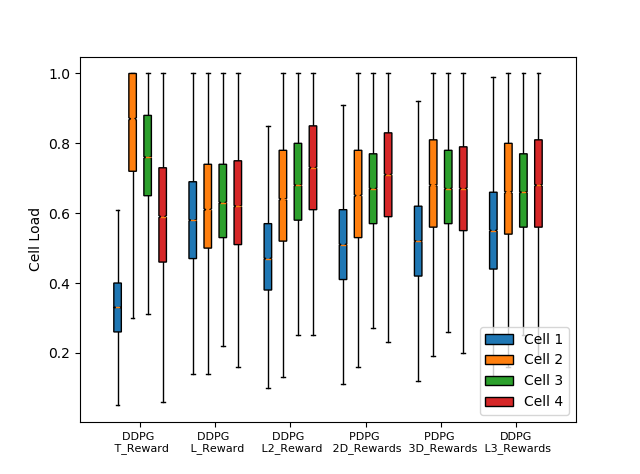}
    \caption{Cell load result of using different RL algorithms}
    \label{cell_load_alg}
\end{figure}

\begin{figure}
    \centering
    \includegraphics[width = 0.5 \linewidth, height = 0.4 \linewidth]{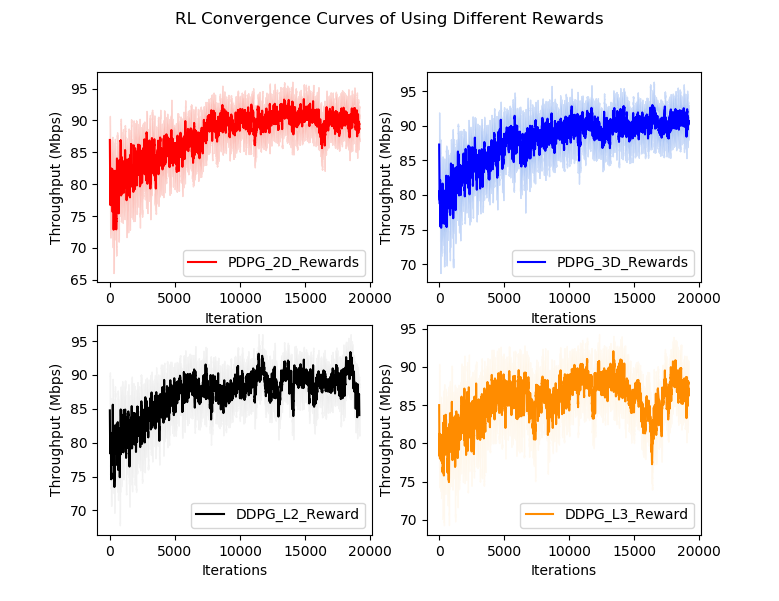}
    \caption{RL convergence curves of using different rewards.}
        \label{convergence_RL}
\end{figure}

In Fig. \ref{cell_load_alg}, we have cell load box plots of the underlying 4 cells, where the corresponding throughput is the same to Fig. \ref{Throughput_CDF_RL_Alg}. As we can see, the scalar reward with only throughput reward has less balanced cell load. Consequently, the unbalanced cell load affects the overall throughput optimization, in which the algorithm can only converge to a local optimum with less performance advantage. Meanwhile, the scalar reward with only maximum load based approach scarifies the throughput gain to favor a more balanced load which also deviates from the joint optimization purpose. Using linear combination of the previous scalar rewards to form another scalar reward can lead to convergence issue as demonstrated in Fig. \ref{convergence_RL}. This is because of the ambiguity issue by mixing scalar rewards as we discussed in Section \ref{RL_form}. The convergence curve together with the throughput and cell load distribution corroborate the advantage of choosing the vector reward based approaches. Overall, the vector reward based RL approaches can achieve more balanced cell load and high throughput which yield the joint optimality on our objective.

\subsection{Comparison to Static Approach}

\begin{figure}
    \centering
    \includegraphics[width = 0.5 \linewidth, height = 0.4 \linewidth]{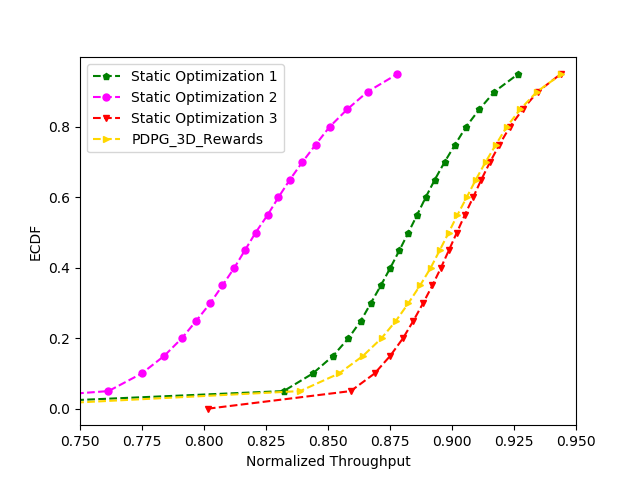}
    \caption{Throughput CDF of static optimization and vector reward based RL algorithm.}
    \label{throughput_alg2}
\end{figure}

\begin{figure}
    \centering
    \includegraphics[width = 0.5 \linewidth, height = 0.4 \linewidth]{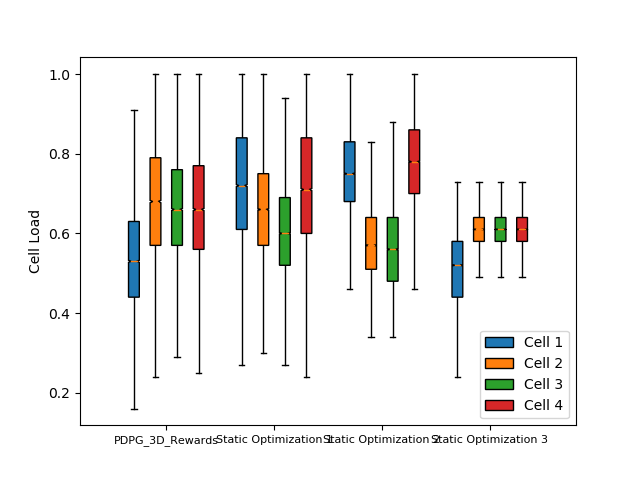}
    \caption{Cell load distribution of static optimization and vector reward based RL algorithm.}
        \label{cell_load_alg2}
\end{figure}

\begin{figure}
    \centering
    \includegraphics[width = 0.5 \linewidth, height = 0.4 \linewidth]{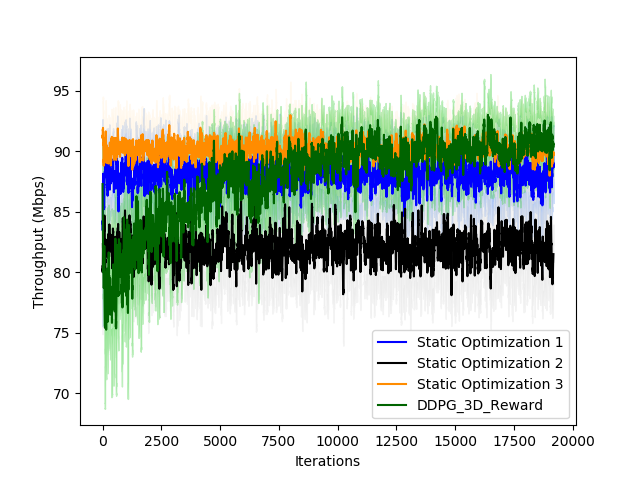}
    \caption{The throughput changing over time for static optimization and vector reward based RL algorithm. }
    \label{convergence_alg2}
\end{figure}

To make a self-contained evaluation, we add static optimization based brute-force algorithms to the comparison, where the details of these algorithms can be found in Appendix. Accordingly, the resulting CDF curves of the network throughput are presented in Fig. \ref{throughput_alg2} (See Algorithm \ref{algorithm_12} and Algorithm \ref{algorithm_13}). Regarding the labels in this figure, the ``static optimization 1'' stands for Algorithm \ref{algorithm_13}; the ``static optimization 2'' stands for Algorithm \ref{algorithm_12}, where the antenna tilt angles are only allowed to be changed per 2 hours in terms of the mobility model for these two algorithms; While the ``static optimization 3'' stands for Algorithm \ref{algorithm_13} using the best cross validated $w$, but the antenna tilt angles can be changed every 15 mins. We can observe that the CDF curve of our introduced RL method is very similar to the static optimization 3. Here, the CDF curve of the RL algorithm is with respect to when the RL algorithm converged (The convergence is considered as when the number of iterations are beyond 10000 as shown in Fig. \ref{convergence_alg2}.). Fig. \ref{cell_load_alg2} shows that the RL approach achieves a balanced cell load as similar as the ideal static case - static optimization 3. The cell load obtained from other static optimization approaches have high variance. Thus, the corresponding throughput performance is deteriorated. Moreover, it is important to note our utilized solvers for the static optimization problem are still sub-optimum approach. The computational complexity of finding the true global optimum for the static optimization problem is prohibitively high. Overall, we believe the RL approach achieves good performance on cell load balancing as well as throughput maximization.

\subsection{Mobility Models}
\label{mobility_model}
To fully evaluate our introduced RL algorithm, we test the algorithm using another mobility model: self-similar least-action walk (SLAW) model \cite{SLAW_model}, where the SLAW model assumes that user mobility is based on clusters (users only move among these clusters). Fig. \ref{convergence_RL} shows the convergence of the vector reward based RL algorithm under the two mobility models. Similarly as our previous findings, using vector reward requires less parameters re-tuning over different mobility models. This is because the vector-formed reward conveys the objective features to the RL value network through a more efficient way. Overall, we conclude that our introduced vector reward based RL algorithm is a robust learning approach with less cross-validation requirements. 

\begin{figure}
    \centering
    \includegraphics[width = 0.5 \linewidth, height = 0.4 \linewidth]{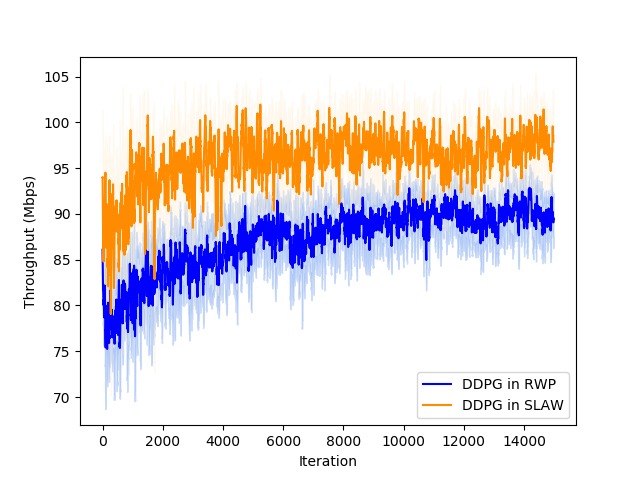}
    \caption{The time dynamic of RL convergence in SLAW model, where metric for the upper one is the network peak load and the lower one is the throughput}
    \label{time_dynamic}
\end{figure}

\section{Conclusion}
\label{sec7}
In this paper, we introduced a vector reward based multiple objective reinforcement learning algorithm. It is utilized to jointly optimize the cell load balance and throughput in mobility management tasks. We choose FD-MIMO antenna tilt and handover CIO adjustment as the RL action space. In addition, the RL reward is designed as a vector, where each entry represents a feature of the joint optimization objective. Accordingly, to promote the Pareto optimality, the introduced RL agent is configured with multiple value networks along with a single policy network. The weighting parameters on the multiple objectives are embedded into the value network which is used for guiding the policy network learning afterward. Moreover, we developed a static formulation for the same joint optimization problem and compare it to our introduced RL algorithm. The algorithm evaluations are presented in different ways including the cell load distribution, throughput distribution and the learning curves of RL algorithms.

For future work, we can consider asynchronous actions based multiple objective learning. Meanwhile, we can include other handover protocols into the framework, such as A2 and A5 events based inter-frequency handover. Moreover, other network features can be incorporated to test the performance of using high dimension vector rewards. 

\section*{Acknowledgement}


\appendix
\section{Comparison: A Static Formulation}
\label{sec5}
We now consider a static formulation of the joint optimization on load balancing and throughput maximization. To do so, we directly drop the expectation and time constraint in (\ref{optimization0}), and set $\gamma = 0$. Therefore, we have the following formulation,
\begin{equation}
\label{optimization2}
\begin{aligned}
    &\max_{{\boldsymbol I}(t), {\boldsymbol b}(t) }  \sum_n R_n(t) + \sum_n F_n(t)\\
    &s.t. \quad  \boldsymbol{I}(t)=\{I_{n, k}(t): I_{n,k}(t) \in \{0, 1\}, n \in N, k \in K\}\\
    &\qquad {\boldsymbol b}(t) = \{b_{n}(t):b_{n}(t) \in \{\theta_0, \theta_1, \cdots, \theta_{M-1} \}, n \in N\},\\
\end{aligned}
\end{equation}

Comparing the above formulation to (\ref{optimization0}), we can notice the following difference and practical limitations,
\begin{itemize}
    \item To solve this static problem, it requires perfect knowledge of all ${p_{n,k}(b_n(t))}$ at every sample time which imposes a large overhead to the user feedback.
    \item Due to the integer constraints, the complexity becomes very high. Moreover, it requires BSs to solve out ${\boldsymbol I}(t)$ and ${\boldsymbol b}(t)$ on every time slot. 
    \item The formulation treats user association as a variable. However, the user association cannot be directly translated to the values of the CIOs in A3-events. Therefore, it is not compatible to the handover operations in current cellular systems.
\end{itemize}
Therefore, we only consider the above formulation as an evaluation approach to our RL algorithm in the simulation. Ideally, the static way can yield the optimal solution at every time. Thus it can serve as an upper bound for the RL algorithm. 

\begin{algorithm}[!t]
    \caption{Relaxed Brute-Force for a small $\lambda$}
    \begin{algorithmic}[1]
        \label{algorithm_12}
    \renewcommand{\algorithmicrequire}{\textbf{Input:}}
    \renewcommand{\algorithmicensure}{\textbf{Output:}}
    \REQUIRE ${\mathcal B} (t)$ (The set of BSs' tilt angles combinations)
    \ENSURE  ${\boldsymbol I}(t)$ and ${\boldsymbol b}(t)$
     \STATE \textit{Initialisation} : ${\mathcal O} = \emptyset$, ${\boldsymbol I}(t) = \emptyset$
     \FOR {${\boldsymbol b}(t) \in {\mathcal B} (t)$}
     \FOR{$\exists I_{n,k}(t) == \emptyset$}
     \FOR{$n \in N$}
     \STATE $I_{n,k}(t) = 1$ and $I_{n',k}(t) = 0$, where $k = \arg max \, \{r_{n,k}(t): I_{n,k}(t) = \emptyset\}$ and $n' \in N/n$ 
     \ENDFOR
     \ENDFOR
     \STATE $\mathcal O = \mathcal O \cup \max_n (L_n({\boldsymbol I}(t), {\boldsymbol b}(t)))$
     \ENDFOR
    \RETURN ${{\boldsymbol I}(t), {\boldsymbol b}(t)} =\arg min \,{\mathcal O}$
    \end{algorithmic} 
\end{algorithm}

\begin{algorithm}[!t]
    \caption{Relaxed Brute-Force for a fair $\lambda$}
    \begin{algorithmic}[1]
    \label{algorithm_13}
    \renewcommand{\algorithmicrequire}{\textbf{Input:}}
    \renewcommand{\algorithmicensure}{\textbf{Output:}}
    \REQUIRE The set of BSs' tilt angles combinations: ${\mathcal B} (t)$, A threshold for association policy decision $w = {\lambda \over {1+\lambda}}$.
    \ENSURE  ${\boldsymbol I}(t)$ and ${\boldsymbol b}(t)$
     \STATE \textit{Initialization} : ${\mathcal O} = \emptyset$, ${\boldsymbol I}(t) = \emptyset$
     \FOR {${\boldsymbol b}(t) \in {\mathcal B} (t)$}
     \FOR{$\exists I_{n,k}(t) == \emptyset$}
     \STATE Draw a random number $c$ from [0,1]
     \IF{$c < w$}
     \FOR{$n \in N$}
     \STATE $I_{n,k}(t) = 1$ and $I_{n',k}(t) = 0$, where $k = \arg max \, \{r_{n,k}(t): I_{n,k}(t) = \emptyset\}$ and $n' \in N/n$ 
     \ENDFOR
     \ELSE
     \STATE Randomly choose $k \in \{k: I_{n, k} = \emptyset\}$
     \STATE $I_{n,k}(t) = 1$ and $I_{n',k}(t) = 0$, where $n = \arg max \, p_{n,k}(b_n(t))$ and $n' \in N/n$ 
     \ENDIF
     \ENDFOR
     \STATE $\mathcal O = \mathcal O \cup U_n({\boldsymbol I}(t), {\boldsymbol b}(t))$
     \ENDFOR
    \RETURN ${{\boldsymbol I}(t), {\boldsymbol b}(t)} =\arg min \,{\mathcal O}$
    \end{algorithmic} 
\end{algorithm}
\subsection{Heuristic Brute-Force Solvers}
Note that (\ref{optimization2}) can be equivalently written as, 
\begin{equation}
\label{optimization12}
\begin{aligned}
    &\max_{{\boldsymbol I}(t), {\boldsymbol b}(t) } \sum_n F_n({\boldsymbol I}(t), {\boldsymbol b}(t))\\
    &s.t. \quad  \boldsymbol{I}(t)=\{I_{n, k}(t): I_{n,k}(t) \in \{0, 1\}, n \in N, k \in K\}\\
    &\qquad {\boldsymbol b}(t) = \{b_{n}(t):b_{n}(t) \in \{\theta_0, \theta_1, \cdots, \theta_{M-1} \}, n \in N\}\\
    &\qquad R_n({\boldsymbol I}(t), {\boldsymbol b}(t)) > \phi, n \in N
\end{aligned}
\end{equation}
where $\phi$ is a parameter to avoid trivial solutions, such as all users are disconnected to BSs (In this case, all cell load are zero). Since we don't have an analytical expression of the objective, brute-force search is considered as our primary approach. However, the number of user association combinations is prohibitively large. To narrow down the searching region, we consider a heuristic approach to approximate
\begin{align}
    \max_{{\boldsymbol I}(t) } \sum_n F_n({\boldsymbol I}(t), {\boldsymbol b}(t))
\end{align} 
for a given ${\boldsymbol b}(t)$. We consider determining the user association in a round-robin manner: Each BS is allocated with an equal number of associated users, where the associated user for each BS is based on the ranking of user rates. Intuitively, this can be considered as a heuristic way to average the throughput $R_n({\boldsymbol I}(t), {\boldsymbol b}(t))$ over all cells. Accordingly, the algorithm is summarized in Algorithm \ref{algorithm_12}. 

Alternatively, the inner loop for user association assignment can be considered as solving 
\begin{align}
    \max_{{\boldsymbol I}(t) } \sum_n R_n({\boldsymbol I}(t), {\boldsymbol b}(t))
\end{align} 
A heuristic approach is assigning users to the BS with maximum transmission power. This association strategy may break the cell load balance but avoid user failure links.  Therefore, we can combine the previous two heuristic association strategies and obtain Algorithm (\ref{algorithm_13}). Particularly, the two user association strategies are mixed through a random binary decision, where the decision threshold is proportional to the weight ratio between the two objectives, i.e., ${\lambda \over {1+\lambda}}$.


%





\ifCLASSOPTIONcaptionsoff
  \newpage
\fi



\bibliographystyle{IEEEtran}
\bibliography{IEEEabrv,./bibtex/bib/reference}

\begin{thebibliography}{10}
\providecommand{\url}[1]{#1}
\csname url@samestyle\endcsname
\providecommand{\newblock}{\relax}
\providecommand{\bibinfo}[2]{#2}
\providecommand{\BIBentrySTDinterwordspacing}{\spaceskip=0pt\relax}
\providecommand{\BIBentryALTinterwordstretchfactor}{4}
\providecommand{\BIBentryALTinterwordspacing}{\spaceskip=\fontdimen2\font plus
\BIBentryALTinterwordstretchfactor\fontdimen3\font minus
  \fontdimen4\font\relax}
\providecommand{\BIBforeignlanguage}[2]{{%
\expandafter\ifx\csname l@#1\endcsname\relax
\typeout{** WARNING: IEEEtran.bst: No hyphenation pattern has been}%
\typeout{** loaded for the language `#1'. Using the pattern for}%
\typeout{** the default language instead.}%
\else
\language=\csname l@#1\endcsname
\fi
#2}}
\providecommand{\BIBdecl}{\relax}
\BIBdecl

\bibitem{r2020}
R.~{Shafin}, L.~{Liu}, V.~{Chandrasekhar}, H.~{Chen}, J.~{Reed}, and J.~C.
  {Zhang}, ``Artificial intelligence-enabled cellular networks: A critical path
  to beyond-5g and 6g,'' \emph{IEEE Wireless Communications}, vol.~27, no.~2,
  pp. 212--217, 2020.

\bibitem{klaine2017survey}
P.~V. Klaine, M.~A. Imran, O.~Onireti, and R.~D. Souza, ``A survey of machine
  learning techniques applied to self-organizing cellular networks,''
  \emph{IEEE Communications Surveys \& Tutorials}, vol.~19, no.~4, pp.
  2392--2431, 2017.

\bibitem{mao17}
Y.~{Mao}, C.~{You}, J.~{Zhang}, K.~{Huang}, and K.~B. {Letaief}, ``A survey on
  mobile edge computing: The communication perspective,'' \emph{IEEE
  Communications Surveys Tutorials}, vol.~19, no.~4, pp. 2322--2358,
  Fourthquarter 2017.

\bibitem{nam2013full}
Y.-H. Nam, B.~L. Ng, K.~Sayana, Y.~Li, J.~Zhang, Y.~Kim, and J.~Lee,
  ``Full-dimension mimo (fd-mimo) for next generation cellular technology,''
  \emph{IEEE Communications Magazine}, vol.~51, no.~6, pp. 172--179, 2013.

\bibitem{zhang17}
H.~{Zhang}, N.~{Liu}, X.~{Chu}, K.~{Long}, A.~{Aghvami}, and V.~C.~M. {Leung},
  ``Network slicing based 5g and future mobile networks: Mobility, resource
  management, and challenges,'' \emph{IEEE Communications Magazine}, vol.~55,
  no.~8, pp. 138--145, Aug 2017.

\bibitem{imran2014challenges}
A.~Imran, A.~Zoha, and A.~Abu-Dayya, ``Challenges in 5g: how to empower son
  with big data for enabling 5g,'' \emph{IEEE network}, vol.~28, no.~6, pp.
  27--33, 2014.

\bibitem{ruiz2015analysis}
J.~M. Ruiz-Aviles, M.~Toril, S.~Luna-Ram{\'\i}rez, V.~Buenestado, and
  M.~Regueira, ``Analysis of limitations of mobility load balancing in a live
  lte system,'' \emph{IEEE wireless communications letters}, vol.~4, no.~4, pp.
  417--420, 2015.

\bibitem{andrews2014overview}
J.~G. Andrews, S.~Singh, Q.~Ye, X.~Lin, and H.~S. Dhillon, ``An overview of
  load balancing in hetnets: Old myths and open problems,'' \emph{IEEE Wireless
  Communications}, vol.~21, no.~2, pp. 18--25, 2014.

\bibitem{ye13}
Q.~{Ye}, B.~{Rong}, Y.~{Chen}, M.~{Al-Shalash}, C.~{Caramanis}, and J.~G.
  {Andrews}, ``User association for load balancing in heterogeneous cellular
  networks,'' \emph{IEEE Transactions on Wireless Communications}, vol.~12,
  no.~6, pp. 2706--2716, June 2013.

\bibitem{damnjanovic2011survey}
A.~Damnjanovic, J.~Montojo, Y.~Wei, T.~Ji, T.~Luo, M.~Vajapeyam, T.~Yoo,
  O.~Song, and D.~Malladi, ``A survey on 3gpp heterogeneous networks,''
  \emph{IEEE Wireless communications}, vol.~18, no.~3, pp. 10--21, 2011.

\bibitem{lopez2012mobility}
D.~Lopez-Perez, I.~Guvenc, and X.~Chu, ``Mobility management challenges in 3gpp
  heterogeneous networks,'' \emph{IEEE Communications Magazine}, vol.~50,
  no.~12, pp. 70--78, 2012.

\bibitem{schrittwieser2019mastering}
J.~Schrittwieser, I.~Antonoglou, T.~Hubert, K.~Simonyan, L.~Sifre, S.~Schmitt,
  A.~Guez, E.~Lockhart, D.~Hassabis, T.~Graepel \emph{et~al.}, ``Mastering
  atari, go, chess and shogi by planning with a learned model,'' \emph{arXiv
  preprint arXiv:1911.08265}, 2019.

\bibitem{mnih2013playing}
V.~Mnih, K.~Kavukcuoglu, D.~Silver, A.~Graves, I.~Antonoglou, D.~Wierstra, and
  M.~Riedmiller, ``Playing atari with deep reinforcement learning,''
  \emph{arXiv preprint arXiv:1312.5602}, 2013.

\bibitem{vinyals2019grandmaster}
O.~Vinyals, I.~Babuschkin, W.~M. Czarnecki, M.~Mathieu, A.~Dudzik, J.~Chung,
  D.~H. Choi, R.~Powell, T.~Ewalds, P.~Georgiev \emph{et~al.}, ``Grandmaster
  level in starcraft ii using multi-agent reinforcement learning,''
  \emph{Nature}, vol. 575, no. 7782, pp. 350--354, 2019.

\bibitem{zhou2017optimizing}
Z.~Zhou, X.~Li, and R.~N. Zare, ``Optimizing chemical reactions with deep
  reinforcement learning,'' \emph{ACS central science}, vol.~3, no.~12, pp.
  1337--1344, 2017.

\bibitem{mnih2014recurrent}
V.~Mnih, N.~Heess, A.~Graves \emph{et~al.}, ``Recurrent models of visual
  attention,'' in \emph{Advances in neural information processing systems},
  2014, pp. 2204--2212.

\bibitem{5grrc}
G.~T. 38.331, ``Nr; radio resource control (rrc); protocol specification.''

\bibitem{5gphysical}
G.~T. 38.213, ``Nr; physical layer procedures for control.''

\bibitem{bethanabhotla2014user}
D.~Bethanabhotla, O.~Y. Bursalioglu, H.~C. Papadopoulos, and G.~Caire, ``User
  association and load balancing for cellular massive mimo,'' in \emph{2014
  Information Theory and Applications Workshop (ITA)}.\hskip 1em plus 0.5em
  minus 0.4em\relax IEEE, 2014, pp. 1--10.

\bibitem{razaviyayn2013linear}
M.~Razaviyayn, M.~Hong, and Z.-Q. Luo, ``Linear transceiver design for a mimo
  interfering broadcast channel achieving max--min fairness,'' \emph{Signal
  Processing}, vol.~93, no.~12, pp. 3327--3340, 2013.

\bibitem{singh2013offloading}
S.~Singh, H.~S. Dhillon, and J.~G. Andrews, ``Offloading in heterogeneous
  networks: Modeling, analysis, and design insights,'' \emph{IEEE Transactions
  on Wireless Communications}, vol.~12, no.~5, pp. 2484--2497, 2013.

\bibitem{Hasan18}
M.~M. {Hasan}, S.~{Kwon}, and J.~{Na}, ``Adaptive mobility load balancing
  algorithm for lte small-cell networks,'' \emph{IEEE Transactions on Wireless
  Communications}, vol.~17, no.~4, pp. 2205--2217, April 2018.

\bibitem{wang2010dynamic}
H.~Wang, L.~Ding, P.~Wu, Z.~Pan, N.~Liu, and X.~You, ``Dynamic load balancing
  and throughput optimization in 3gpp lte networks,'' in \emph{Proceedings of
  the 6th international wireless communications and mobile computing
  conference}, 2010, pp. 939--943.

\bibitem{son2009dynamic}
K.~Son, S.~Chong, and G.~De~Veciana, ``Dynamic association for load balancing
  and interference avoidance in multi-cell networks,'' \emph{IEEE Transactions
  on Wireless Communications}, vol.~8, no.~7, pp. 3566--3576, 2009.

\bibitem{ao2017approximation}
W.~C. Ao and K.~Psounis, ``Approximation algorithms for online user association
  in multi-tier multi-cell mobile networks,'' \emph{IEEE/ACM Transactions on
  Networking}, vol.~25, no.~4, pp. 2361--2374, 2017.

\bibitem{mwanje2013q}
S.~S. Mwanje and A.~Mitschele-Thiel, ``A q-learning strategy for lte mobility
  load balancing,'' in \emph{2013 IEEE 24th Annual International Symposium on
  Personal, Indoor, and Mobile Radio Communications (PIMRC)}.\hskip 1em plus
  0.5em minus 0.4em\relax IEEE, 2013, pp. 2154--2158.

\bibitem{mwanje2016cognitive}
S.~S. Mwanje, L.~C. Schmelz, and A.~Mitschele-Thiel, ``Cognitive cellular
  networks: A q-learning framework for self-organizing networks,'' \emph{IEEE
  Transactions on Network and Service Management}, vol.~13, no.~1, pp. 85--98,
  2016.

\bibitem{kudo2014q}
T.~Kudo and T.~Ohtsuki, ``Q-learning based cell selection for ue outage
  reduction in heterogeneous networks,'' in \emph{2014 IEEE 80th Vehicular
  Technology Conference (VTC2014-Fall)}.\hskip 1em plus 0.5em minus 0.4em\relax
  IEEE, 2014, pp. 1--5.

\bibitem{xu2019deep}
Y.~Xu, W.~Xu, Z.~Wang, J.~Lin, and S.~Cui, ``Deep reinforcement learning based
  mobility load balancing under multiple behavior policies,'' in \emph{ICC
  2019-2019 IEEE International Conference on Communications (ICC)}.\hskip 1em
  plus 0.5em minus 0.4em\relax IEEE, 2019, pp. 1--6.

\bibitem{xu2019load}
------, ``Load balancing for ultradense networks: A deep reinforcement
  learning-based approach,'' \emph{IEEE Internet of Things Journal}, vol.~6,
  no.~6, pp. 9399--9412, 2019.

\bibitem{wang2018handover}
Z.~Wang, L.~Li, Y.~Xu, H.~Tian, and S.~Cui, ``Handover control in wireless
  systems via asynchronous multiuser deep reinforcement learning,'' \emph{IEEE
  Internet of Things Journal}, vol.~5, no.~6, pp. 4296--4307, 2018.

\bibitem{zappone2018user}
A.~Zappone, L.~Sanguinetti, and M.~Debbah, ``User association and load
  balancing for massive mimo through deep learning,'' in \emph{2018 52nd
  Asilomar Conference on Signals, Systems, and Computers}.\hskip 1em plus 0.5em
  minus 0.4em\relax IEEE, 2018, pp. 1262--1266.

\bibitem{lillicrap2015continuous}
T.~P. Lillicrap, J.~J. Hunt, A.~Pritzel, N.~Heess, T.~Erez, Y.~Tassa,
  D.~Silver, and D.~Wierstra, ``Continuous control with deep reinforcement
  learning,'' \emph{arXiv preprint arXiv:1509.02971}, 2015.

\bibitem{van2014multi}
K.~Van~Moffaert and A.~Now{\'e}, ``Multi-objective reinforcement learning using
  sets of pareto dominating policies,'' \emph{The Journal of Machine Learning
  Research}, vol.~15, no.~1, pp. 3483--3512, 2014.

\bibitem{miettinen2002scalarizing}
K.~Miettinen and M.~M. M{\"a}kel{\"a}, ``On scalarizing functions in
  multiobjective optimization,'' \emph{OR spectrum}, vol.~24, no.~2, pp.
  193--213, 2002.

\bibitem{van2017hybrid}
H.~Van~Seijen, M.~Fatemi, J.~Romoff, R.~Laroche, T.~Barnes, and J.~Tsang,
  ``Hybrid reward architecture for reinforcement learning,'' in \emph{Advances
  in Neural Information Processing Systems}, 2017, pp. 5392--5402.

\bibitem{kaelbling1996reinforcement}
L.~P. Kaelbling, M.~L. Littman, and A.~W. Moore, ``Reinforcement learning: A
  survey,'' \emph{Journal of artificial intelligence research}, vol.~4, pp.
  237--285, 1996.

\bibitem{sutton1988learning}
R.~S. Sutton, ``Learning to predict by the methods of temporal differences,''
  \emph{Machine learning}, vol.~3, no.~1, pp. 9--44, 1988.

\bibitem{munos2016safe}
R.~Munos, T.~Stepleton, A.~Harutyunyan, and M.~Bellemare, ``Safe and efficient
  off-policy reinforcement learning,'' in \emph{Advances in Neural Information
  Processing Systems}, 2016, pp. 1054--1062.

\bibitem{SLAW_model}
K.~{Lee}, S.~{Hong}, S.~J. {Kim}, I.~{Rhee}, and S.~{Chong}, ``Slaw: A new
  mobility model for human walks,'' in \emph{IEEE INFOCOM 2009}, 2009, pp.
  855--863.

\end{thebibliography}
\end{document}